%% file: camera_ready.tex
\documentclass[letterpaper]{article} 
\usepackage{aaai25}  
\usepackage{times}  
\usepackage{helvet}  
\usepackage{courier}  
\usepackage[hyphens]{url}  
\usepackage{graphicx} 
\usepackage{natbib}  
\usepackage{caption} 
\usepackage{algorithm}
\usepackage{algorithmic}

\usepackage{multirow}
\usepackage{booktabs}
\usepackage{amssymb}
\usepackage{newfloat}
\usepackage{listings}
\DeclareCaptionStyle{ruled}{labelfont=normalfont,labelsep=colon,strut=off} 
\floatstyle{ruled}
\newfloat{listing}{tb}{lst}{}
\floatname{listing}{Listing}
\title{HOGSA: Bimanual Hand-Object Interaction Understanding with 3D Gaussian Splatting Based Data Augmentation}
\author{
    Wentian Qu\textsuperscript{\rm 1,\rm 2},
    Jiahe Li\textsuperscript{\rm 1,\rm 2},
    Jian Cheng\textsuperscript{\rm 1,\rm 2},
    Jian Shi\textsuperscript{\rm 3,\rm 2},
    Chenyu Meng\textsuperscript{\rm 1,\rm 2},\\
    Cuixia Ma\textsuperscript{\rm 1,\rm 2},
    Hongan Wang\textsuperscript{\rm 1,\rm 2},
    Xiaoming Deng\textsuperscript{\rm 1,\rm 2}\thanks{indicates corresponding author.},
    Yinda Zhang\textsuperscript{\rm 4}\footnotemark[1]
}
\affiliations{
   \textsuperscript{\rm 1}Institute of Software, Chinese Academy of Sciences,
    \textsuperscript{\rm 2}University of Chinese Academy of Sciences,\\
    \textsuperscript{\rm 3}Institute of Automation, Chinese Academy of Sciences,
    \textsuperscript{\rm 4}Google\\

    \{wentian2019, lijiahe2021, chengjian, cuixia, hongan,xiaoming\}@iscas.ac.cn,\\ 
    jian.shi@ia.ac.cn,
    mengchenyu21@mails.ucas.ac.cn, 
    yindaz@google.com
}

\usepackage{bibentry}

\begin{document}

\maketitle

\input{0_abstract}
\input{1_introduction}

\input{2_related_work}

\input{3_method}

\input{4_experiment}

\input{5_conclusion}

\input{7_acknowledgments}
\bibliography{aaai25}
\clearpage
\input{6_supp}
\end{document}

%% file: 0_abstract.tex
\begin{abstract}
Understanding of bimanual hand-object interaction plays an important role in robotics and virtual reality. However, due to significant occlusions between hands and object as well as the high degree-of-freedom motions, it is challenging to collect and annotate a high-quality, large-scale dataset, which prevents further improvement of bimanual hand-object interaction-related baselines. In this work, we propose a new 3D Gaussian Splatting based data augmentation framework for bimanual hand-object interaction, which is capable of augmenting existing dataset to large-scale photorealistic data with various hand-object pose and viewpoints. First, we use mesh-based 3DGS to model objects and hands, and to deal with the rendering blur problem due to multi-resolution input images used, we design a super-resolution module. Second, we extend the single hand grasping pose optimization module for the bimanual hand object to generate various poses of bimanual hand-object interaction, which can significantly expand the pose distribution of the dataset. Third, we conduct an analysis for the impact of different aspects of the proposed data augmentation on the understanding of the bimanual hand-object interaction. We perform our data augmentation on two benchmarks, H2O and Arctic, and verify that our method can improve the performance of the baselines.
\end{abstract}

\begin{links}
    \link{Project Page}{https://iscas3dv.github.io/HOGSA/}
\end{links}

%% file: 1_introduction.tex
\section{1. Introduction}
Understanding of the bimanual hand-object interaction~\cite{fan2023arctic, kwon2021h2o}, especially the estimation of the pose of the hand-object and the contact relationship, plays an increasingly important role in robotics and virtual reality applications. 
One of the most popular approaches to address this problem is deep learning-based methods, which require large-scale bimanual hand-object interaction dataset with rich annotations. 
However, due to significant occlusions and high-degree-of-freedom motions of bimanual hand-object interaction, it is still challenging to collect and annotate a high-quality dataset, which prevents further improvement of the task.

\input{teaser}

To address the challenges of data scarcity and inaccurate 3D annotations, existing work has explored data augmentation methods with synthetic data under conventional rendering pipelines
~\cite{corona2020ganhand, jian2023affordpose, yang2022artiboost}. However, these approaches usually require complex and time-consuming 3D scanning and post-processing to capture high-quality shapes and texture maps of the hand and object, and additional blending weights of hand models is necessary for augmentation, which requires substantial expertise~\cite{romero2022embodied,deng2021hand}. Moreover, capturing a realistic texture map (i.e. subtle details and natural appearance of the hand and object) from observed images is difficult, thus it often results in rendering results of hand and object that lack realism ~\cite{qian2020html}. 
Recently, benefitting from the scene representation ability, neural rendering methods such as NeRF~\cite{mildenhall2021nerf} and 3DGS~\cite{kerbl20233d} enable high-quality data augmentation by synthesizing novel views~\cite{feldmann2024nerfmentation} or novel hand poses~\cite{qu2023novel}. NeRFmentation~\cite{feldmann2024nerfmentation} uses NeRF to perform data augmentation for the monocular depth estimation task in static scenes, but it cannot break the accuracy bottleneck when the scene changes significantly. HO-NeRF~\cite{qu2023novel} builds a pose-driven NeRF for hand-object interaction scenarios and demonstrates the potential to generate diverse data, yet it requires offline modeling of hands and objects and a time-costing rendering process, making it infeasible for data augmentation of large-scale dataset. 
Although these neural rendering methods can support realistic novel view synthesis \cite{wang2021neus} and are potentially useful for data augmentation, they still suffer from image blur due to multi-resolution image input and inaccurate annotations~\cite{yu2024mip,barron2021mip}. Unrealistic images cannot resolve the gap between real images and synthetic images, which will lead to a degradation of model performance.
Another key factor that affects the baseline performance is the diversity of poses in the dataset~\cite{deng2021hand}. Therefore, it is necessary to establish a data augmentation approach of bimanual hand-object interaction that enables efficient rendering, various feasible hand-object poses, and photorealistic rendering images.  

In this paper, we propose a 3DGS-based data augmentation framework \textbf{H}and-\textbf{O}bject \textbf{G}aussian \textbf{S}platting \textbf{A}ugmentation (\textbf{HOGSA}) for bimanual hand-object interaction understanding. First, we use mesh-based 3DGS to model the hand and object based on the hand-object interaction images, which can efficiently synthesize interaction images with the input hand-object pose and viewpoints. Second, in order to enhance the pose diversity of the dataset, we use the pose optimization module to generate diverse poses of two hands and object to drive the hand-object Gaussian splatting model to render images of novel interaction poses. Third, in order to ensure the realism of the rendered images, we design  the super-resolution module to improve the rendering quality of the coarse images generated by 3DGS. Finally, we combine our augmented dataset with the original dataset to refine the baseline of bimanual hand-object interaction, and conduct a systematic analysis of different aspects that affect interaction understanding accuracy in the augmented dataset.
We evaluate our method on two main benchmarks H2O~\cite{kwon2021h2o} and Arctic~\cite{fan2023arctic}, and the baseline performances are improved with our augmented dataset.

The contributions of our method can be summarized as follows: 1) A 3DGS-based data augmentation framework for bimanual hand-object interaction understanding; 2) a super-resolution module and a pose optimization module to improve the realism and pose diversity in the data augmentation; 3) We provide fine-tuning models using our HOGSA that achieve state-of-the-art results on H2O and Arctic benchmarks, and make a systematic analysis of the impact for augmented data on accuracy.

%% file: teaser.tex
\begin{figure}[t]
\centering%
\includegraphics[width=\linewidth]{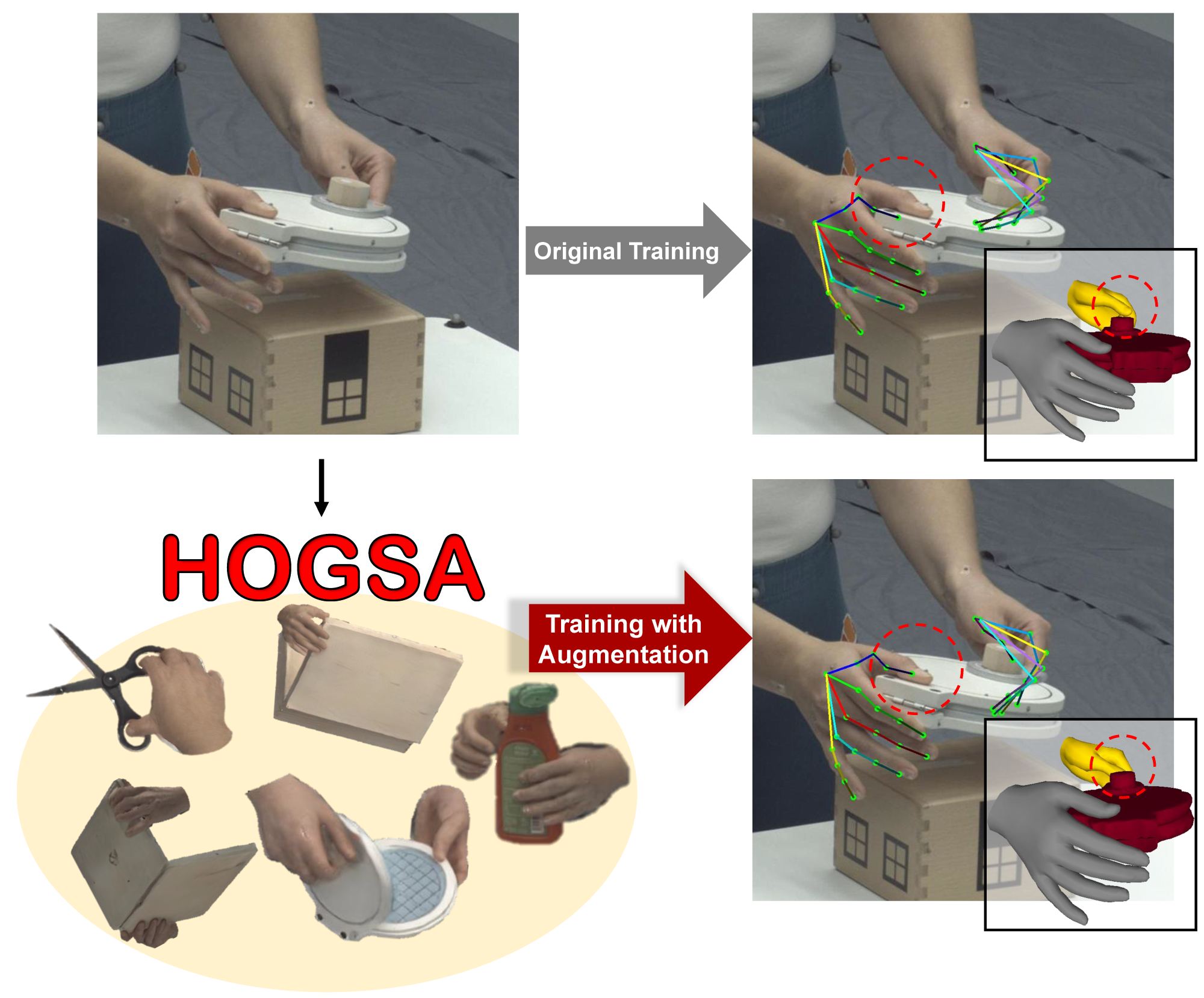}
\caption{We propose a new 3DGS-based data augmentation framework for bimanual hand-object interaction to augment existing dataset with various hand-object pose and viewpoints. Our method can improve the performance of the baselines, and achieve more accurate pose and contact.}
\label{fig:teaser}
\end{figure}

%% file: 2_related_work.tex
\section{2. Related Work}
\input{pipeline}
\noindent \textbf{Hand-Object Interaction Understanding.}
Hand-object interaction understanding focuses on 3D reconstruction, pose estimation and contact estimation.
Early approaches mainly rely on predefined templates to estimate hand and object poses~\cite{cao2021reconstructing, corona2020ganhand, liu2021semi, tekin2019h+, yang2024learning, qi2024hoisdf}. These template-based methods face significant challenges in realism gap with real-world. Recent advancements have transitioned towards template-free approaches that leverage large-scale 3D hand-object interaction datasets \cite{chen2023gsdf, ye2022s, fan2024hold, zhang2024moho}. Nonetheless, the limited diversity and quantity of 3D data restricts models' ability to generalize effectively across various scenarios.
Several datasets with hand-object contact annotations have advanced contact estimation~\cite{taheri2020grab, brahmbhatt2020contactpose, fan2023arctic}. Contactopt \cite{grady2021contactopt} leverages the predicted contact points to optimize grasping. Some studies \cite{narasimhaswamy2020detecting, shan2020understanding} infer 2D bounding boxes of hands with contact from RGB images, while others \cite{rogez2015understanding, fan2023arctic} explore 3D contact inference. However, these data-driven approaches are suffer from the diverse poses and viewpoints in dataset. 
In this paper, we tackle with the limitation of data collection, and propose a 3DGS-based data augmentation framework to enhance the performance of the baselines.

\noindent \textbf{Data Augmentation for Hand-Object Interaction.}
Data augmentation is essential for enhancing model performance in hand-object interaction understanding. Existing methods can be divided into rendering-based and generative-based methods.
Rendering-based methods generate plausible hand-object interaction images by designing specific pipelines \cite{corona2020ganhand, jian2023affordpose, yang2022artiboost,gao2022dart,li2023chord} or utilizing off-the-shelf tools such as GraspTTA \cite{jiang2021hand} and UniDexGrasp \cite{xu2023unidexgrasp}. However, the rendering-based methods often lack realism.
Generative-based methods will create more realistic hand-object interaction images. HOGAN \cite{hu2022hand} synthesizes novel views using target poses as guidance. Several approaches employ conditional diffusion models to generate hand grasping. Affordance Diffusion \cite{ye2023affordance} generates hand-object interaction images, conditioned on a hand orientation mask. HandBooster \cite{xu2024handbooster} and HOIDiffusion~\cite{zhang2024hoidiffusion} synthesizes realistic hand-object images with diverse appearances, poses, views, and backgrounds. 
MANUS~\cite{pokhariya2023manus} utilizes 3DGS to model hand and object respectively and combine them to form a data set.
However, the augmentation approaches are mostly focus on interacting between single hand and object. These setups are more serious mutual occlusion when migrating to bimanual hand-object interacting tasks, making the model difficult to learn effective pose priors.

In this work, to achieve high realism of augumented data, we enforce the generated bimanual hand-object interaction results to meet geometric constraints and overcome the realism issues of 3DGS, which enables our data augmentation method to improve the performance of the baseline.

%% file: pipeline.tex
\begin{figure*}[ht]
\centering%
\includegraphics[width=\linewidth]{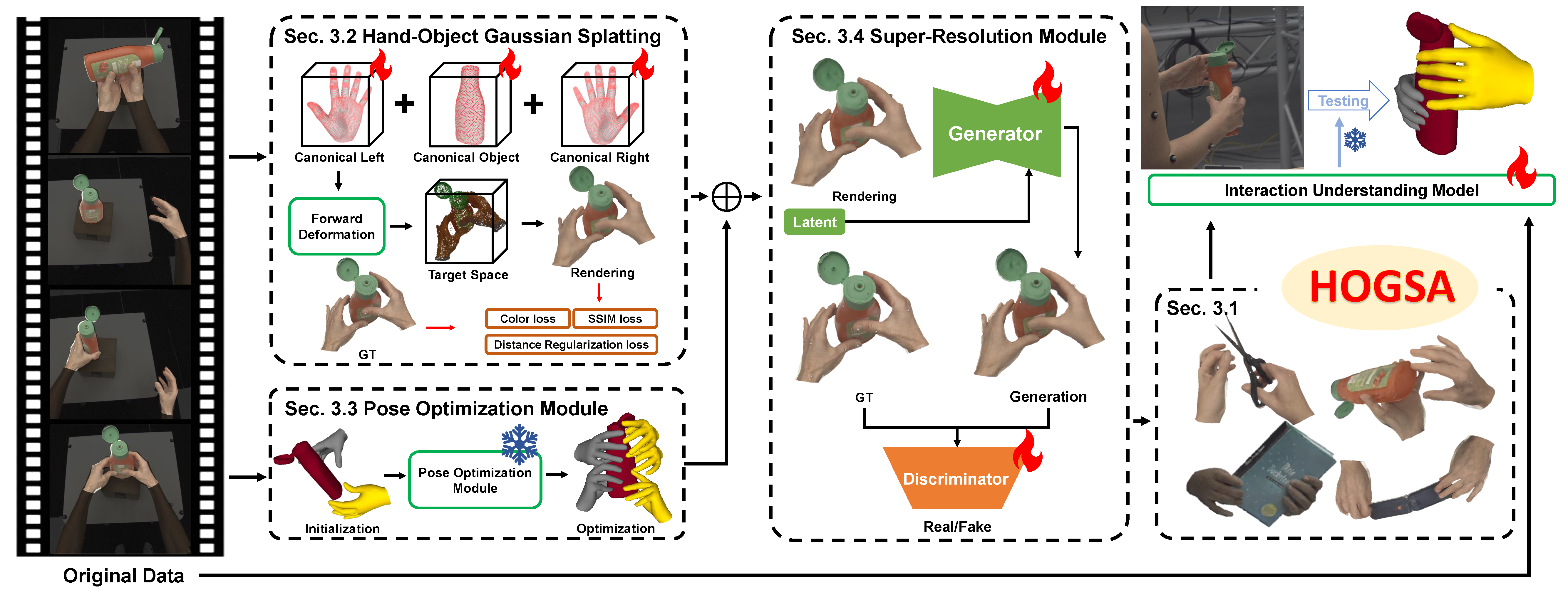}
\caption{Overview of our data augmentation framework for bimanual hand-object interaction. Based on the original dataset, we first establish mesh-based 3DGS models and input the original poses to pose optimization module to expand the diversity of interaction. The novel pose and 3DGS can be combined to render the low-quality image, which is then fed into the super-resolution module to further enhance the realism. Based on the above modules, we can automatically build an expanded dataset and support model fine-tuning for the interaction understanding baseline to improve performance.
}
\label{fig:pipeline}
\end{figure*}

%% file: 3_method.tex
\section{3. Method}
In this work, we propose a data augmentation framework for bimanual hand-object interaction to improve the accuracy of the baselines for interaction pose estimation (Fig.~\ref{fig:pipeline}). 
Our data augmentation method HOGSA contains diverse poses and realistic rendering images, and can achieve end-to-end generation based on novel hand-object and camera poses.
We introduce the pipeline of our data augmentation in Section~3.1 and detail the implementation in the resting sections.
We first use the mesh-based 3DGS method to model the left hand, right hand and object models based on the input hand-object interaction images respectively (Sec.~3.2). Then we use the Pose Optimization Module (POM)  (Sec.~3.3) to optimize the novel poses of the bimanual hand-object interaction, ensuring the realism of the grasp while expanding the diversity of the pose. Combining Hand-Object Gaussian Splatting (HOGS) and Pose Optimization Module, we render low-quality images, and design the Super-Resolution Module (SRM) (Sec.~3.4) to improve the rendering quality.  Finally, we use both the augmentation and the original dataset to refine the baseline to improve the accuracy of the interaction pose estimation task.

\input{hogs_aug}
\subsection{3.1 Hand-Object Gaussian Splatting Augmentation}
Based on the original dataset, we first build a HOGS model for each sequence.
Second, based on the original poses, we use POM to optimize and expand the diversity of poses of the bimanual hand-object interaction. Then we use the HOGS model with these novel poses of hand-object interaction and camera poses to render new images as our new data augmentation process, and feed the new image into SRM to improve the image quality. The above three modules provide an image rendering pipeline, pose diversity, and rendering realism, thus achieving an efficient data augmentation solution. 
We use the bimanual hand-object interaction dataset Arctic~\cite{fan2023arctic} and H2O~\cite{kwon2021h2o} to generate synthetic data HOGSA respectively, and the results are shown in Fig.~\ref{fig:hogs_aug}. To organize the data used to train the baseline, we are inspired by GANerated Hands~\cite{mueller2018ganerated} and add COCO2017~\cite{lin2014microsoft} images as background to fuse the generated image, and crop it to $224\times224$ resolution (as shown in Fig.~\ref{fig:hogs_aug_with_bg}). Finally, we combine HOGSA and the original data to improve the performance of the baseline.

\input{hogs_aug_with_bg}

\subsection{3.2 Hand-Object Gaussian Splatting}
Our method utilizes color images as input to represent the complex dynamic scene using mesh-based 3D Gaussian Splatting, accomplishing rendering under novel poses and novel views. For initialization, we define the Gaussian kernel on MANO-HD~\cite{chen2023hand} template hand mesh and object mesh, respectively, which are named canonical space. The standard 3D Gaussian Splatting~\cite{kerbl20233d} can be defined as:
\begin{equation}
    \mathbf{G}(\mathbf{x})=e^{-(\mathbf{x}-\mathbf{x}_{c})^{\top} \Sigma^{-1}(\mathbf{x}-\mathbf{x}_{c})},
\end{equation}
where $\mathbf{x}_{c}\in\mathbb{R}^{3}$ represents the Gaussian center and $\Sigma\in\mathbb{R}^{3\times3}$ represents the covariance matrix, which is parameterized by rotation matrix $\mathbf{R}$ and scaling matrix $\mathbf{S}$ as $\Sigma=\mathbf{R}\mathbf{S}\mathbf{S}^{\top}\mathbf{R}^{\top}$. 
Inspired by GaMeS~\cite{waczynska2024games}, we convert the Gaussian kernels into mesh surfaces and the Gaussian center can be defined as:
\begin{equation}
    \mathbf{x}_{c}=\mathbf{\beta}\mathbf{V}=\beta_{1}\mathbf{v}_{1}+\beta_{2}\mathbf{v}_{2}+\beta_{3}\mathbf{v}_{3},
\end{equation}
where $\mathbf{\beta}=\{\beta_{i}\}_{i=1}^{3}$ are the trainable parameters that satisfies $\beta_{1}+\beta_{2}+\beta_{3}=1$ and $\mathbf{V}=\{\mathbf{v}_{i}\}_{i=1}^{3}$ are the vertices from the mesh face. For the definition of the rotation matrix $\mathbf{R}=\{\mathbf{r}_{i}\}_{i=1}^{3}$, $\mathbf{r}_{1}$ is the surface normal, $\mathbf{r}_{2}$ is the vector from the center of the surface $\mathbf{m}=mean(\mathbf{v}_{1},\mathbf{v}_{2},\mathbf{v}_{3})$ to the vertex $\mathbf{v}_{1}$, and $\mathbf{r}_{3}$ is obtained by orthonormalizing the vector for $\mathbf{r}_{1},\mathbf{r}_{2}$. For the scaling matrix $\mathbf{S}=\{s_{i}\}_{i=1}^{3}$, we define $s_{1}=s_{2}=\Vert \mathbf{m}-\mathbf{v}_{2}\Vert,s_{3}=\langle \mathbf{v}_{2},\mathbf{R}_{3}\rangle$.

\noindent \textbf{Hand Model.} To model the hand motion, we use the blending weights from MANO-HD to initialize the hand Gaussian skinning weights $\mathbf{w}$, where $\mathbf{w}=(w_{1}, w_{2},...,w_{n})$ with $n$ joints. For the target frame, we use bone transformation matrix $\mathbf{B}$ and blending weights to deform the Gaussian kernels in canonical space to the target space as:
\begin{equation}
\mathbf{x}_{t}^{h}=(\sum_{i=1}^{n}{w_{i}\mathbf{B}_{i}})\cdot\mathbf{x}_{c}^{h},
\end{equation}
where $\mathbf{x}_{c}^{h}$ is the hand canonical space point, $\mathbf{x}_{t}^{h}$ is the corresponding point in target space.

\noindent \textbf{Object Model.} We use 6D rigid pose $\mathbf{T}\in SE(3)$ to perform rigid transformation from canonical space to target space:
\begin{equation}
\mathbf{x}_{t}^{o}=\mathbf{T} \times \mathbf{x}_{c}^{o},
\end{equation}
where $\mathbf{x}_{c}^{o}$ is the object canonical space point, $\mathbf{x}_{t}^{o}$ represents the mapping point in target space.

\noindent \textbf{Alpha Blending.} The 3D Gaussian points in target space are projected to 2D Gaussian, then sorted based on depth and the color value of the pixel can be calculated by:
\begin{equation}
\mathbf{c}=\sum_{i=1}^{N}{\mathbf{c}_{i}\alpha_{i}}\prod_{j=1}^{i-1}{(1-\alpha_{j})},
\end{equation}
where $\mathbf{c}$ represents the color value of the pixel, $\mathbf{c}_{i}$ and $\alpha_{i}$ represent the color and pixel translucency of the $i$-th Gaussian kernel respectively. 

\noindent \textbf{Training.} For the Gaussian kernels defined in left-hand, right-hand and object model, the optimized parameters include mesh vertex positions $\mathbf{V}$, patch weights $\alpha$, scale factor $s$, spherical harmonics coefficients and opacity. 
The loss function of our HOGS model can be expressed as:
\begin{equation}
L_{HOGS}=(1-\lambda_{SSIM})L_{1}+\lambda_{SSIM}L_{SSIM}+\lambda_{R}L_{R},
\end{equation}
where L1 loss $L_{1}$ and SSIM loss $L_{SSIM}$ are used to minimizes the difference between the input and rendered images, the distance regularization loss $L_{R}$ constrains the vertices in the canonical space not to deviate far from the initial mesh, $\lambda_{SSIM}, \lambda_{R}$ are hyperparameters.

We split the datasets for training and testing our HOGS. For Arctic, we select subjects except '$s03$' and '$s05$' to build the HOGS model to meet the data division of the interaction understanding task. We select the sequences '$grab\_01$' and '$use\_01$' to train HOGS, and cropped the original images to a resolution of $1400\times1000$. For H2O, we select subjects except ‘$subject4$’ to build the HOGS model. We select sequences except ‘$o2$’ scene for training HOGS and keep the original image resolution $1280\times720$. 

\subsection{3.3 Pose Optimization Module}

In order to ensure the diversity of pose in augmentation, we need to synthesize more bimanual hand-object interaction results that meet geometric constraints. Inspired by GraspTTA~\cite{jiang2021hand} with a fitting strategy, we extend their solution of single-hand grasping to bimanual hand object interaction. We discard the 'GraspCVAE' model used to generate the initial hand pose and only fixed the parameters of the 'ContactNet' model for contact map inference. We randomly perturb the hands extracted from the original dataset a small distance away from the object as the initial pose. Based on the point cloud of the initial pose, a contact map $\Omega$ can be calculated. At the same time, the point cloud can be input into 'ContactNet' to predict a contact map $\hat{\Omega}$ under prior knowledge. We leverage a self-supervised consistency loss $L_{C}=\Vert \Omega - \hat{\Omega} \Vert^{2}_{2}$ to enforce a reasonable contact. We also use hand-centric loss $L_{H}$ and penetration loss $L_{P}$ to ensure the physically plausible interaction. The loss function is formulated as follows:
\begin{equation}
L_{POM}=\sum_{i\in\{l,r\}}{(\lambda_{C}L_{C}^{i}+\lambda_{H}L_{H}^{i}+\lambda_{P}L_{P}^{i})},
\end{equation}
where $l$ and $r$ represent the left and right hands, $\lambda_{C}, \lambda_{H}, \lambda_{P}$ are hyperparameters. For an initial input, we fix the object pose and optimize the pose of both hands separately with 200 iterations to get the optimized pose. Then we input the optimized pose and camera parameters into HOGS model to synthesize a bimanual hand-object interaction image.

For the initial input pose, we generate a transformation matrix with random rotation between $[0,20^{\circ}]$ around the x, y, and z axis and apply the rotation to the original pose. We then perturb the distance by which the hands are farther away from the object. We calculate the distance and direction of the hand root joint relative to the object, then translate it $5\%$ of the relative distance away from the object, and then add a perturbation of $[0,6cm]$ to the position of the hand and object, respectively. Note that for the Arctic dataset, we impose an angle of $[0.01\pi, 0.2\pi]$ on the one-dimensional rotation of the articulated object.

\subsection{3.4 Super-Resolution Module}
We notice that the different resolutions of the images and the deviation in pose annotation to train 3DGS can result in blurry rendered images with artifacts to train 3DGS . 
Therefore, we use CNN to improve the rendering quality in 3DGS. Inspired by StyleAvatar~\cite{wang2023styleavatar}, we can use the encoder-decoder backbone to learn the local and global features of the image and integrate it into the GAN framework to improve the rendering quality. We input the coarse image $\mathbf{I}_{C}$ rendered by 3DGS into 'StyleUNet' to obtain the refined image $\mathbf{I}_{R}$ and use the ground truth image to constrain its generation quality. The loss function is defined as follows:
\begin{equation}
L_{SRM}=\lambda_{1}L_{1}+\lambda_{VGG}L_{VGG}+L_{GAN},
\end{equation}
where $L_{1}$ represents L1 loss, $L_{VGG}$ represents VGG loss, $L_{GAN}$ represents the loss used in adversarial learning, and $\lambda_{1}, \lambda_{VGG}$ are hyperparameters. 
In this module, we follow the data split used in HOGS. We leverage HOGS to render images based on the training set and then pair them with the ground truth to train the SRM model.

%% file: hogs_aug.tex
\begin{figure}[ht]
\centering%
\includegraphics[width=\linewidth]{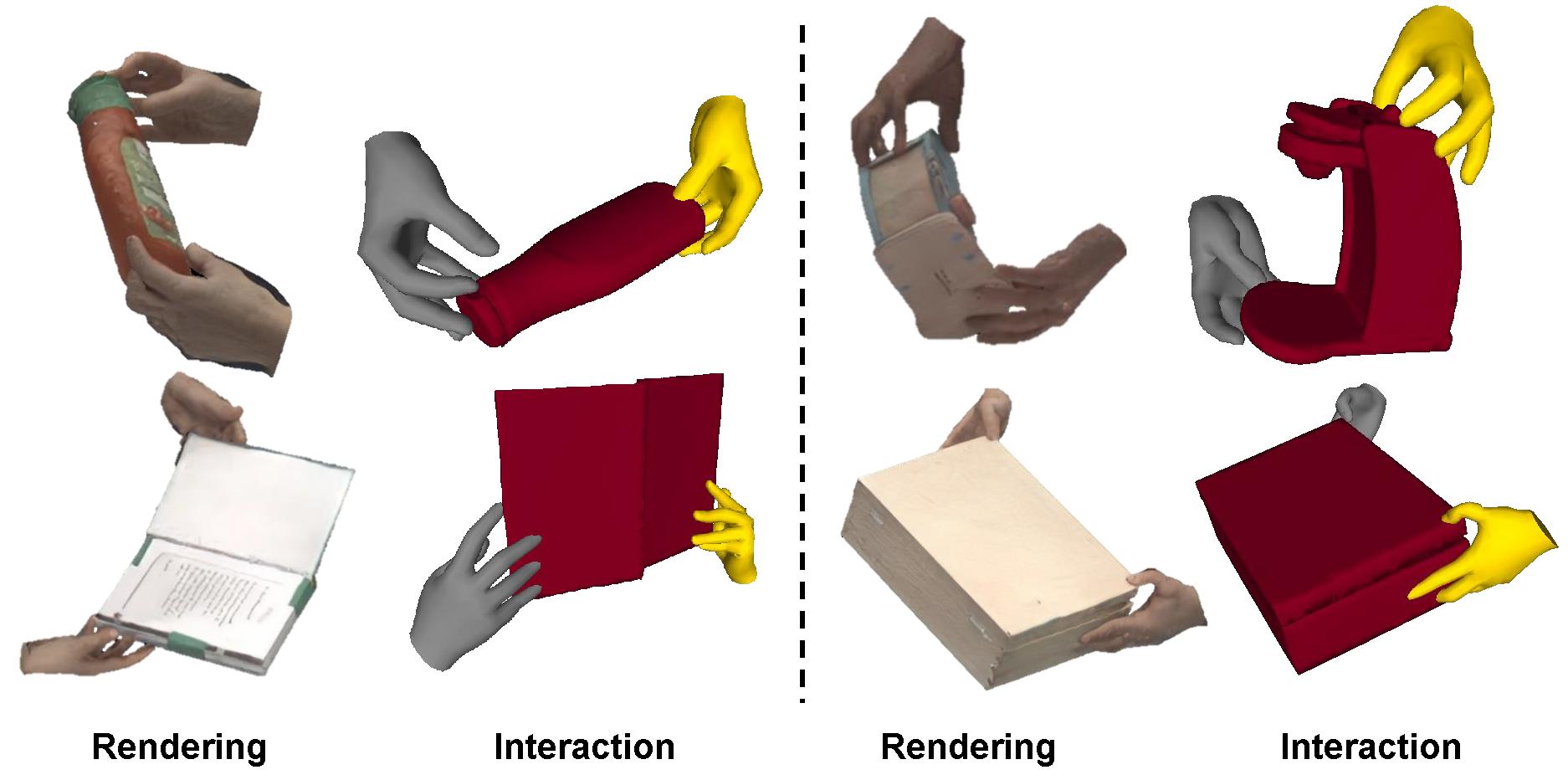}
\caption{Examples of our HOGSA, which contains diverse interactive poses and ensures the realism of the images. 
}
\label{fig:hogs_aug}
\end{figure}

%% file: hogs_aug_with_bg.tex
\begin{figure}[ht]
\centering%
\includegraphics[width=\linewidth]{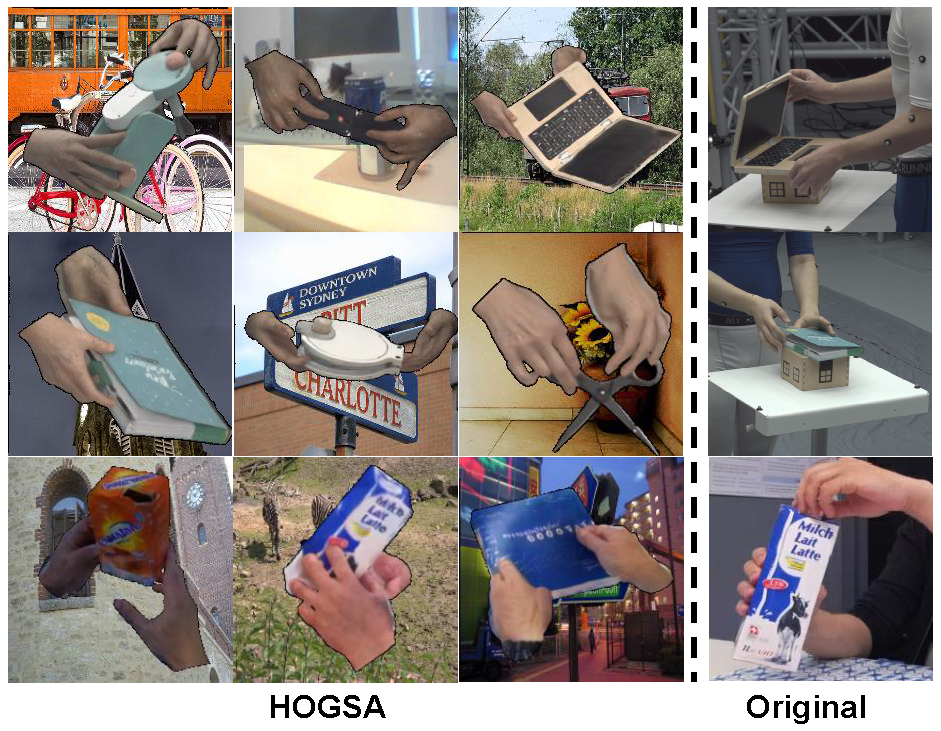}
\caption{The augmented data we used to train the baseline. Compared with the original data, our images ensure realism and have various poses.
}
\label{fig:hogs_aug_with_bg}
\end{figure}

%% file: 4_experiment.tex
\input{fig_comparison}
\input{table_comparison}

\section{4. Experiment}

\subsection{4.1 Datasets, Baseline and Evaluation Metric}

\noindent \textbf{Datasets.}
We evaluate our method on Arctic~\cite{fan2023arctic} and H2O~\cite{kwon2021h2o}.
Arctic is a dataset for dexterous bimanual hand-object interaction. It consists of 10 humans, with manipulating 11 articulated objects. We follow allocentric validation split in Arctic to train and test the baseline.
H2O is a dataset for two hands manipulating objects. It consists of 4 subjects, with 6 scenarios manipulating 8 rigid objects. We select the first 3 subjects for augmentation and baseline training, and use the last subject for testing.

\noindent \textbf{Augmentation.}
In the construction of HOGSA for each benchmark, we consider the interaction of the same subject with the same object as a sequence. For the Arctic dataset, we build a total of 82 HOGS models and generate 1.7M images using the novel poses. Compared to the original 1.5M training images, we have automatically expanded the data by almost $113\%$.
For the original H2O dataset with 0.4M training data, we generate 0.7M images with 24 HOGS models, which has automatically expanded the data by almost $175\%$.

\noindent \textbf{Baseline.} \emph{Interaction Understanding}: We follow Arctic~\cite{fan2023arctic} to define two tasks including 'Consistent Motion Reconstruction' and 'Interaction Field Estimation', and select 'ArcticNet-SF' and 'InterField-SF' in Arctic as the baselines, respectively.
\emph{Augmentation}: We compare with HOIDiffusion~\cite{zhang2024hoidiffusion} ('Arctic+HOID') and replace the mesh-based 3DGS in the HOGS module with the point cloud based 3DGS~\cite{kerbl20233d} ('Arctic+3DGS').

\noindent \textbf{Evaluation Metric.}
To evaluate the quality of the bimanual hand-object interaction understanding, we follow Arctic~\cite{fan2023arctic} to use Contact Deviation (CDev, $mm$), Motion Deviation (MDev, $mm$) and Acceleration Error (ACC, $m/s^{2}$) to measure the accurate hand-object contact, stable move, and smooth motion, respectively. We use the mean per-joint position error (MPJPE, $mm$), the average articulation error (AAE, $n^\circ$), the success rate (SR, $n\%$) to measure the pose accuracy, use the mean relative-root position error (MRRPE, $mm$) to calculate hand-object relative translations and use average distance error to measure the interaction field.
To evaluate the quality of novel view synthesis in ablation study, we follow 3DGS~\cite{kerbl20233d} to adopt PSNR, SSIM and LIPIPS as metrics.

\noindent \textbf{Implementation Details}
Our method is trained on a Nvidia RTX4090 GPU. We train HOGS and SRM modules from scratch. 
We train HOGS with 50,000 iterations, which costs an average of 10 GB of memory.
We train SRM with a total of 150,000 iterations and cost 8 GB of memory. 
For the POM, each initial input needs to be iterated 200 times.
We set hyperparameters $\lambda_{SSIM}, \lambda_{R}, \lambda_{C}, \lambda_{H}, \lambda_{P}, \lambda_{1}, \lambda_{VGG}$ to 0.2, 0.5, 1, 1, 17, 5, 0.03 respectively.

\subsection{4.2 Comparison with Baseline}
\noindent \textbf{Consistent Motion Reconstruction.}
We evaluate this task over the baseline 'ArcticNet-SF' on Arctic and H2O benchmarks, and the results are shown in Tab.~\ref{tab:comparison}. After refining the baseline by adding our expanded dataset to the original training set, we find that all metrics are improved. 
HOIDiffusion deals with the data augmentation for right hand-object interacting scene, and some augmentation images are ambiguous, which will make the baseline difficult to learn effective pose priors under severe occlusion caused by the interaction between two hands and objects, and slightly improves the baseline. 
Our method ensures the diversity of bimanual hand-object pose as well as realism of rendered images, and can effectively improve the performance of baseline. 
Our method takes 0.06s to generate an augmented image, which is also more efficient than HOIDiffusion’s 4.5s.
We show qualitative results in Fig.~\ref{fig:comparison} and find that after adding our data augmentation, the estimation of novel poses and contacts near occlusions are more accurate. 
This is because our augmented dataset provides richer viewpoint and pose priors.
Compared with the original point cloud based 3DGS, our mesh-based 3DGS avoid losing valid information due to points straying too far from the instance based on mesh constraints, which results in higher rendering quality. The original 3DGS rendered images have serious artifacts, which reduces the effect of data augmentation.

\input{inter_field}
\noindent \textbf{Interaction Field Estimation.}
The interaction field estimation measures the relative spatial relations between hands and the object. We show the results using the 'InterField-SF' in Fig.~\ref{fig:comparison}~(b) and Tab.~\ref{tab:inter_field}. After adding augmentation, the baseline is less affected by occlusion, and the rich pose prior makes its estimation more accurate.
The baseline refined by HOGSA promotes stable mutual movement between the hand and the object and effectively reduces pose jitter. Based on the various pose and viewpoints of our augmentation, the occluded right hand position can predict a more accurate contact area without noise (Fig.~\ref{fig:comparison}~(b)). Therefore, our method can be applied to the baseline of various interaction understanding tasks to improve performance.

\input{fig_rendering}
\input{table_rendering}
\subsection{4.3 Ablation Study}
\input{abl_pose}

\input{tsne}

\noindent \textbf{Effect of Super-Resolution Module.}
We compare the rendering quality of images generated by HOGS with and without SRM. The results are shown in Fig.~\ref{fig:rendering} and Tab.~\ref{tab:rendering}. 
The rendering quality is significantly improved after adding SRM, especially in the texture details of objects and the wrinkles on the hands. The local semantic feature information learned by CNN compensates for artifacts and blur caused by images of different resolutions and pose bias. 
It is noted that the SSIM is comparable to the method without SRM. The reason is that the geometry of the image is mainly affected by HOGS, while SRM mainly improves the texture quality of the image.
We compare the impact of augmented datasets with different rendering qualities on the accuracy of consistent motion reconstruction, and the results are shown in the first and third rows of Tab.~\ref{tab:abl_pose}. The SRM further reduces the gap between real and synthetic data, and more realistic images can better improve the performance of the baseline.

\noindent \textbf{Effect of Pose Optimization Module.}
We compare the pose diversity in the original data and the data enhanced by POM. We encode the hand joints and then use the T-SNE clustering~\cite{van2008tsne} to show in Fig.~\ref{fig:tsne}. After adding the POM module, 
the orange points distributed around the blue points complement the diversity of the entire data pose distribution, which shows a reinforcement of the original data.
In order to evaluate the effect of POM, we remove the POM to generate augmented data (w/o POM) and combine them with  original data to train the baseline, and the results are shown in the second and third rows of Tab.~\ref{tab:abl_pose}. We find that while perspective augmentation improves the performance of the model, the prediction accuracy of novel poses is still not significant. Since the POM can enhance the distribution of poses, it can improve the interaction understanding ability of the model.

%% file: fig_comparison.tex
\begin{figure*}[ht]
\centering%
\includegraphics[width=\linewidth]{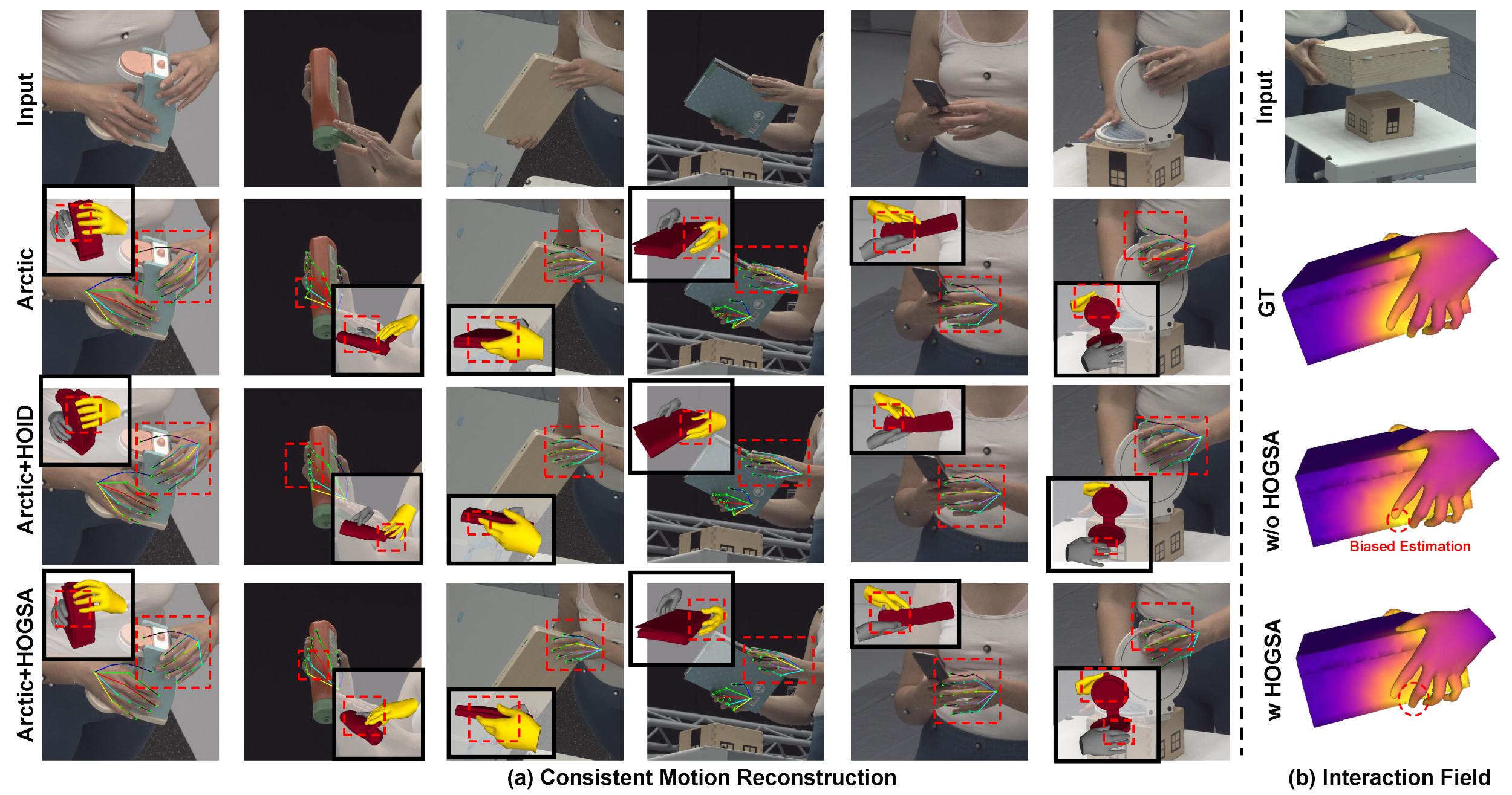}
\caption{Qualitative results of our data augmentation method HOGSA on the baseline. After optimization, the model can cover a wider range of interactive poses and achieve a more accurate estimation of the pose and contact area.
}
\label{fig:comparison}
\end{figure*}

%% file: table_comparison.tex
\begin{table*}[t]
\centering
\fontsize{9}{11}\selectfont
\begin{tabular}{c|c|cc|cc|c|cc}
\hline
 & & \multicolumn{2}{|c|}{Contact and Relative Position}   & \multicolumn{2}{|c|}{Motion}          & \multicolumn{1}{|c|}{Hand}  & \multicolumn{2}{c}{Object} \\\hline
Method & Dataset & CDev$_{ho}$ $\downarrow$& MRRPE$_{rl/ro}$  $\downarrow$ & MDev$_{ho}$  $\downarrow$ & ACC$_{h/o}$ $\downarrow$ & MPJPE$_h$ $\downarrow$ & AAE  $\downarrow$ & SR$\uparrow$\\\hline
\multirow{5}{*}{ArcticNet-SF}  
& Arctic &41.35&50.14/37.59&10.46&6.63/8.80&23.01&5.85& 71.77 \\
& Arctic+HOID&39.19&46.57/36.93&9.75&\textbf{6.36}/8.38&22.44&5.88& 73.56 \\
& Arctic+3DGS&36.07&45.17/34.68&8.89&6.41/7.59&21.78&5.76& 77.06 \\
& Arctic+HOGSA &\textbf{35.23}&\textbf{43.69}/\textbf{33.48}&\textbf{8.77}&\textbf{6.36}/\textbf{7.54}&\textbf{20.96}&\textbf{5.67}&\textbf{77.85} \\
\cline{2-9}
& H2O&35.74&57.65/47.53&5.31&3.80/6.05&34.38&-& 39.80 \\
& H2O+HOGSA&\textbf{32.27}&\textbf{54.63}/\textbf{43.93}&\textbf{5.16}&\textbf{3.79}/\textbf{6.02}&\textbf{32.24}&-& \textbf{45.27} \\
\hline

\end{tabular}
\caption{
Quantitative results of our data augmentation method on the baseline. Our data augmentation method can be automatically applied to different datasets and support improving the performance of different baselines.
}
\label{tab:comparison}
\end{table*}

%% file: inter_field.tex
\begin{table}[t]
\centering
\fontsize{9}{11}\selectfont
\begin{tabular}{c|c|c|c}
\hline
Dataset & HOGSA & Average Distance Error $\downarrow$&ACC $\downarrow$\\
\hline
\multirow{2}{*}{Arctic} & w/o &9.63/9.91&3.01/2.95 \\
                        & w&\textbf{9.30}/\textbf{8.98}&\textbf{2.98}/\textbf{2.79}\\
\hline
\multirow{2}{*}{H2O} & w/o  &7.75/10.86&1.84/1.89 \\
                        & w &\textbf{7.54}/\textbf{9.87}&\textbf{1.82}/\textbf{1.82}\\
\hline
\end{tabular}
\caption{
After the interaction field estimation baseline is fine-tuned using HOGSA, the performance is improved. 
}
\label{tab:inter_field}
\end{table}

%% file: fig_rendering.tex
\begin{figure*}[ht]
\centering%
\includegraphics[width=\linewidth]{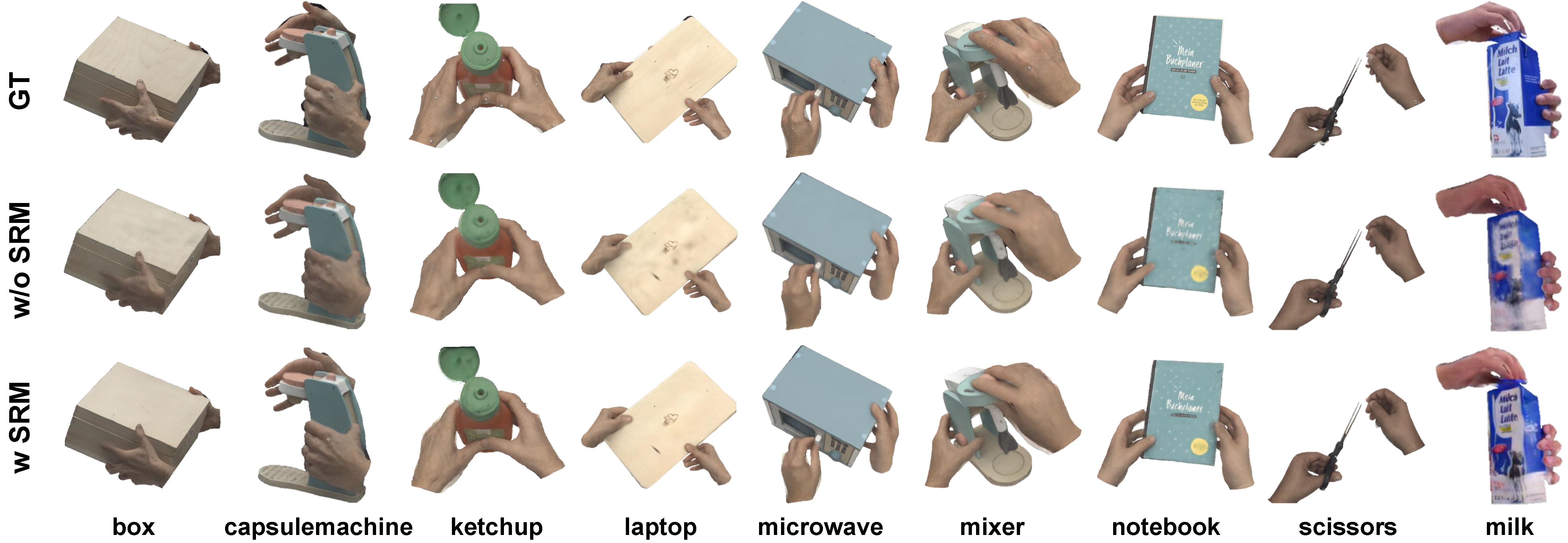}
\caption{Ablation study on SRM. After using SRM, the realism of the image rendered by HOGS is significantly improved, especially the texture details of the object. This greatly reduces the gap between synthetic and real data.
}
\label{fig:rendering}
\end{figure*}

%% file: table_rendering.tex
\begin{table}[t]
\centering
\fontsize{9}{11}\selectfont
\begin{tabular}{c|c|ccc}
\hline
Dataset & Method & PSNR $\uparrow$&SSIM $\uparrow$& LPIPS $\downarrow$\\
\hline
\multirow{2}{*}{Arctic} & w/o SRM &35.07&\textbf{0.988}&0.0194 \\
                        & w SRM&\textbf{36.53}&\textbf{0.988}&\textbf{0.0163}\\
\hline
\multirow{2}{*}{H2O} & w/o SRM &32.32&\textbf{0.987}&0.0133
 \\
                    & w SRM&\textbf{32.44}&0.986&\textbf{0.0117}
\\
\hline
\end{tabular}
\caption{
Our SRM explores the combination of 3DGS and CNN which aggregates pixel and image local semantic information to further improve the realism of image.
}
\label{tab:rendering}
\end{table}

%% file: abl_pose.tex
\begin{table*}[t]
\centering
\fontsize{9}{11}\selectfont
\begin{tabular}{c|c|cc|cc|c|cc}
\hline
                         &               & \multicolumn{2}{|c|}{Contact and Relative Position}   & \multicolumn{2}{|c|}{Motion}           & \multicolumn{1}{|c|}{Hand}  & \multicolumn{2}{c}{Object} \\\hline
Dataset                   & Method        & CDev$_{ho}$ $\downarrow$& MRRPE$_{rl/ro}$  $\downarrow$ & MDev$_{ho}$  $\downarrow$ & ACC$_{h/o}$ $\downarrow$ & MPJPE$_h$ $\downarrow$ & AAE  $\downarrow$ & SR$\uparrow$\\\hline
\multirow{3}{*}{Arctic}  & w/o SRM&36.85&45.40/{34.97}&9.39&6.51/8.05&21.75&5.93& 75.56 \\
                         & w/o POM&37.67&45.35/35.78&9.22&6.42/7.90&21.95&6.08& 75.10 \\
                         & Full&\textbf{35.23}&\textbf{43.69}/\textbf{33.48}&\textbf{8.77}&\textbf{6.36}/\textbf{7.54}&\textbf{20.96}&\textbf{5.67}&\textbf{77.85} \\
\hline
\end{tabular}
\caption{Ablation study on our SRM and POM used for motion reconstruction. We evaluate our method using 'ArcticNet-SF' baseline on Arctic dataset, and the expanded dataset can effectively improve the accuracy of the baseline.
}
\label{tab:abl_pose}
\end{table*}

%% file: tsne.tex
\begin{figure}[ht]
\centering%
\includegraphics[width=\linewidth]{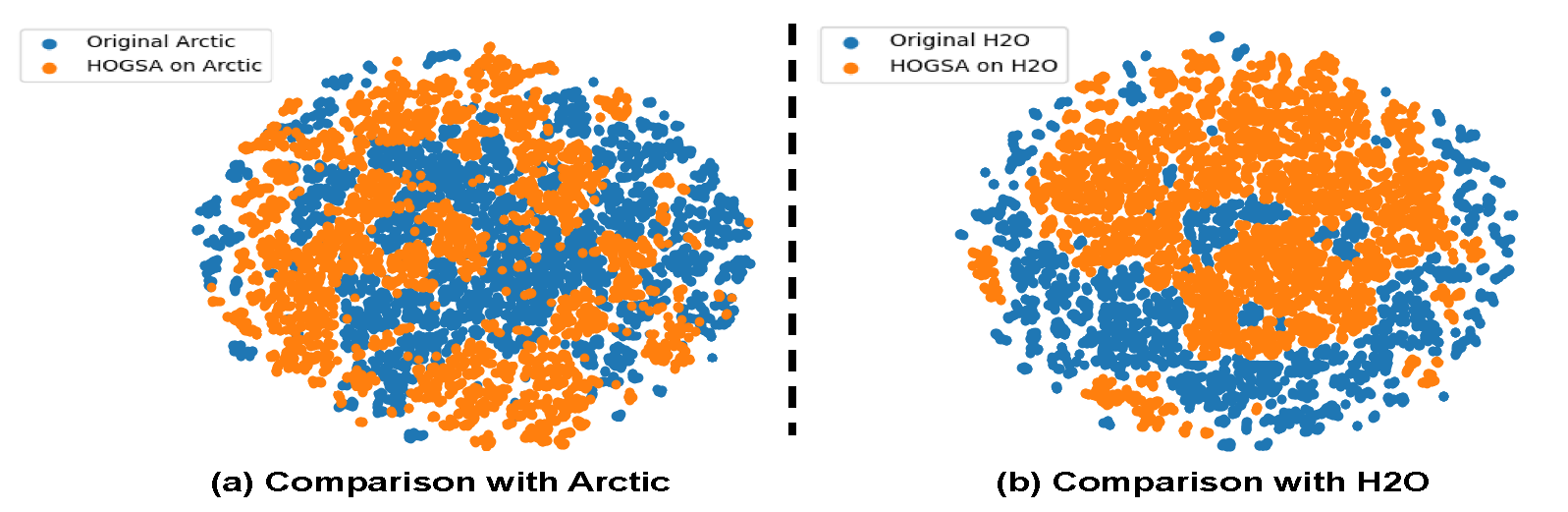}
\caption{We used the joints of both hands to perform T-SNE and evaluate the distribution of poses before and after data augmentation. The poses of our augmented dataset can further enrich the diversity of the original data.
}
\label{fig:tsne}
\end{figure}

%% file: 5_conclusion.tex
\section{5. Conclusion} 
In this work, we propose a new 3DGS-based data augmentation framework for bimanual hand-object interaction, which can augment the existing benchmark to large-scale photorealistic data with various hand-object pose and viewpoints. The expanded dataset can further improve the performance of the existing baseline. We propose the pose optimization module to generate various physically feasible poses of bimanual hand-object interaction and the super-resolution module to improve the realism of rendered images using 3DGS.
We perform our data augmentation on two benchmarks, H2O and Arctic, and verify that our method can improve the performance of the baselines.

%% file: 7_acknowledgments.tex
\section{Acknowledgments}
This work was supported in part by 
National Science and Technology Major Project (2022ZD0119404), National Natural Science Foundation of China (62473356,62373061), Beijing Natural Science Foundation (L232028), CAS Major Project (RCJJ-145-24-14), Science and Technology Innovation Key R\&D Program of Chongqing (CSTB2023TIAD-STX0027), and Beijing Hospitals Authority Clinical Medicine Development of Special Funding Support No. ZLRK202330.

%% file: 6_supp.tex
\appendix
\section{Appendix}
In this supplementary material, we first introduce the implementation details of our method (Sec.~1) and show additional comparison results (Sec.~2). Then we introduce more ablation results and application (Sec.~3).

\section{1. Method Details}

\subsection{1.1 Hand-Object Gaussian Splatting}
\noindent \textbf{Definition of Gaussian Kernel.} We define Gaussian kernels on the mesh faces of the hand and the object, respectively (as shown in the Fig.~\ref{fig:supp_kernel}). For vertices $\mathbf{V}=\{\mathbf{v}_{i}\}_{i=1}^{3}$ from the mesh face, the position of the Gaussian kernel $\mathbf{x}_{c}$ is initialized by defining beta $\mathbf{\beta}=\{\beta_{i}\}_{i=1}^{3}$ (Fig.~\ref{fig:supp_kernel}~(a)). Note that multiple Gaussian kernels can be defined on the same face by defining multiple sets of beta $\beta^{k}$ (Fig.~\ref{fig:supp_kernel}~(b)). In our implementation, we set $k$ = 2, and the number of Gaussian kernels will remain constant during training.

\input{supp_kernel}
\noindent \textbf{Implementation of Distance Regularization Loss.} Our mesh-based 3DGS defines Gaussian kernels on the mesh faces in the canonical space. We optimize the mesh vertices during training, and we define the distance regularization loss $L_{R}$ to constrain the vertices not to deviate too far from the initial mesh to support building geometric details of the instance. The distance regularization loss can be define as:
\begin{equation}
    L_{R} = \frac{1}{N_{v}}\sum_{i=1}^{N_{v}}dist(\mathbf{v}^{i}, \mathbf{M}),
\end{equation}
where $N_{v}$ is the number of vertices in mesh, $\mathbf{v}^{i}$ is the positions of hand or object vertices in canonical space, and $\mathbf{M}$ is the hand or object initial template mesh.
\input{supp_field}

\input{supp_pom_res}

\subsection{1.2 Pose Optimization Module}
In POM, we extract 'ContactNet' from GraspTTA~\cite{jiang2021hand} and extend it from single-hand grasping to bimanual hand-object interaction. This fitting stage can optimize to obtain diverse  interactive poses that meet geometric constraints based on different initialization. We show the diverse interaction results obtained by POM for different objects in Fig.~\ref{fig:supp_pom_res}. During the optimization process, the parameters of the MANO hand model~\cite{romero2022embodied} are optimized, including root rotation and translation, shape parameters, and pose parameters. In order to ensure an efficient optimization process, we adjust the learning rate for pose and translation from the initial $6.25\times10^{-6}$ to $5\times10^{-4}$ and $3\times10^{-5}$, respectively.

\section{2. Additional Comparison Experiments}

\subsection{2.1 Implementation Details of Baseline}
We refer Arctic~\cite{fan2023arctic} to verify the improvement of our data augmentation method HOGSA on the interaction understanding baseline including consistent motion reconstruction and interaction field estimation. During training, we use the pre-trained model provided by the official as the initial model parameters.
We follow official training strategy for 'allocentric' and iterate 15 epoch for each model fine-tuning.

\noindent \textbf{Consistent Motion Reconstruction.} 
In the evaluation on Arctic  dataset~\cite{fan2023arctic}, 
we extract 'ArcticNet-SF' from Arctic~\cite{fan2023arctic} as the baseline, and then initialize the model parameters using the official pre-trained model. 
As shown in Tab.~1 in main paper, we fuse Arctic original dataset and our augmentation dataset as 'Arctic+HOGSA' to fine-tune the model parameter. 
In order to evaluate the effectiveness  using the mesh-based 3DGS method for data augmentation compared to the original 3DGS, we replace HOGS in our data augmentation framework with the original 3DGS. In the original 3DGS method, we define the initial points as the vertices of the mesh in the canonical space and use the same training strategy and loss function. The SRM module is retrained and then combined with the same POM module to achieve data augmentation. We combine the data augmentation results of the original 3DGS with Arctic original data named 'Arctic+3DGS' to fine-tune the baseline.

In the evaluation on H2O  dataset~\cite{kwon2021h2o}, we use the official model parameters pre-trained on Arctic dataset as initialization and refine them on H2O original dataset~\cite{kwon2021h2o}, named 'H2O'. At the same time, we fuse the H2O original data and our expanded dataset to fine-tune the model, named 'H2O+HOGSA'.

\noindent \textbf{Interaction Field Estimation.}
We extract 'InterField-SF' from Arctic~\cite{fan2023arctic} as the baseline, and also initialize the model with the official pre-trained model. We fine-tune the model on the H2O and Arctic original datasets with and without our data augmentation method. 

\subsection{2.2 Comparison Results}

\noindent \textbf{Interaction Field Estimation.}
We show more qualitative results in Fig.~\ref{fig:supp_field}. We find that after fine-tuning the baseline with our expanded dataset, the contact of the occluded part can estimate an accurate interaction field, as shown in the highlighting of the finger contact position (red dotted circle). It benefits from our POM, which ensures the diversity of the pose distribution of our data augmentation, thereby improving the performance of the baseline.

\noindent \textbf{Consistent Motion Reconstruction.}
We show more qualitative results in Fig.~\ref{fig:supp_pose}, including pose estimation and reconstruction results. After being fine-tuned on our data augmentation, the baseline obtains more accurate pose and reconstruction results, especially in the occluded part, the penetration between hand and object is reduced due to the diversity of poses and viewpoints in our dataset, which enables the model to learn rich prior knowledge.

\noindent \textbf{Comparison of Rendering Quality.} We compare the rendering quality of original 3DGS~\cite{kerbl20233d} and our mesh-based HOGS and show the qualitative results in Fig.~\ref{fig:supp_srm}. We find that mesh-based methods can preserve sharper geometric information and more realistic texture details. This is because the mesh-based method can avoid the information loss and quality degradation caused by the Gaussian kernel being too far away from the mesh.

\section{3. Additional Ablation and Application}

\subsection{3.1 Effect of Super-Resolution Module}
Our SRM module is used to further improve the rendering quality of 3DGS by adding semantic feature information of the image. We perform SRM for the original 3DGS and our HOGS, respectively, and the quantitative results are shown in Fig.~\ref{fig:supp_srm}. We find that SRM can be effectively combined with 3DGS to solve the artifacts and blur caused by multi-resolution images training, whether for original 3DGS or mesh-based 3DGS, and it only needs to build one model for multiple instances.

\subsection{3.2 Preliminary Applications}
Our approach also supports applications such as novel view synthesis and pose transfer.

\noindent \textbf{Novel View Synthesis.} We can change the input parameters to generate new perspectives. We show the qualitative results in Fig.~\ref{fig:supp_app} (a). Our method can render 360° around the interaction center, thus ensuring the diversity of perspectives when expanding data.

\noindent \textbf{Pose Transfer.}
Based on the input of the user's interaction pose, we can transfer it to other subject hand models to achieve style transfer. We show the results in Fig.~\ref{fig:supp_app} (b). The image after our pose transfer still ensures the realism of the texture.

\input{supp_pose}
\input{supp_srm}
\input{supp_app}

%% file: supp_kernel.tex
\begin{figure}[ht]
\centering%
\includegraphics[width=\linewidth]{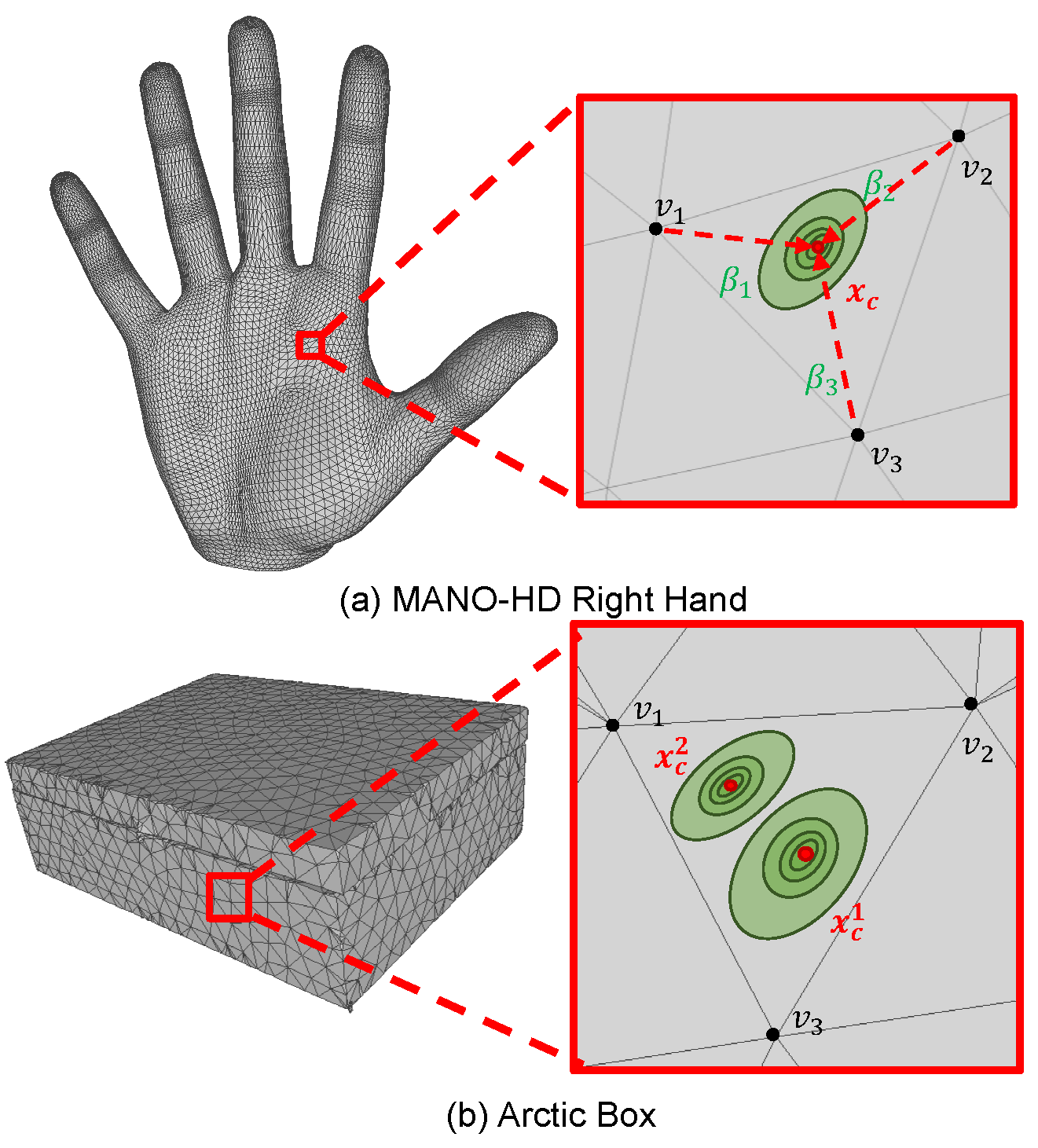}

\caption{Gaussian kernel defined on mesh-based 3DGS. 
}
\label{fig:supp_kernel}
\end{figure}

%% file: supp_field.tex
\begin{figure*}[ht]
\centering%
\includegraphics[width=\linewidth]{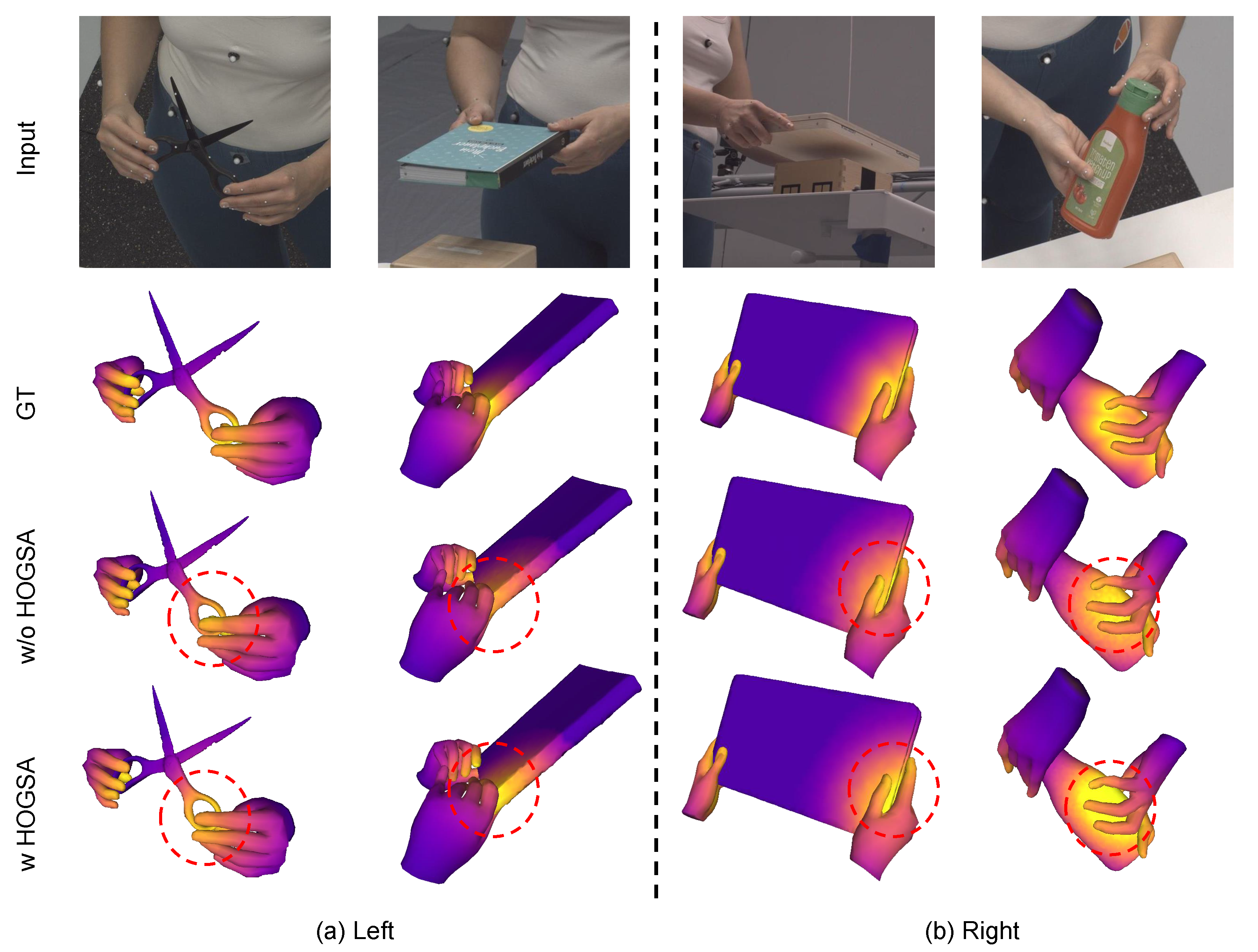}

\caption{Comparison of interaction field estimation. We find that after we optimize the baseline with our data augmentation method, the model predicts more accurate interaction field for occluded parts.
}
\label{fig:supp_field}
\end{figure*}

%% file: supp_pom_res.tex
\begin{figure}[ht]
\centering%
\includegraphics[width=\linewidth]{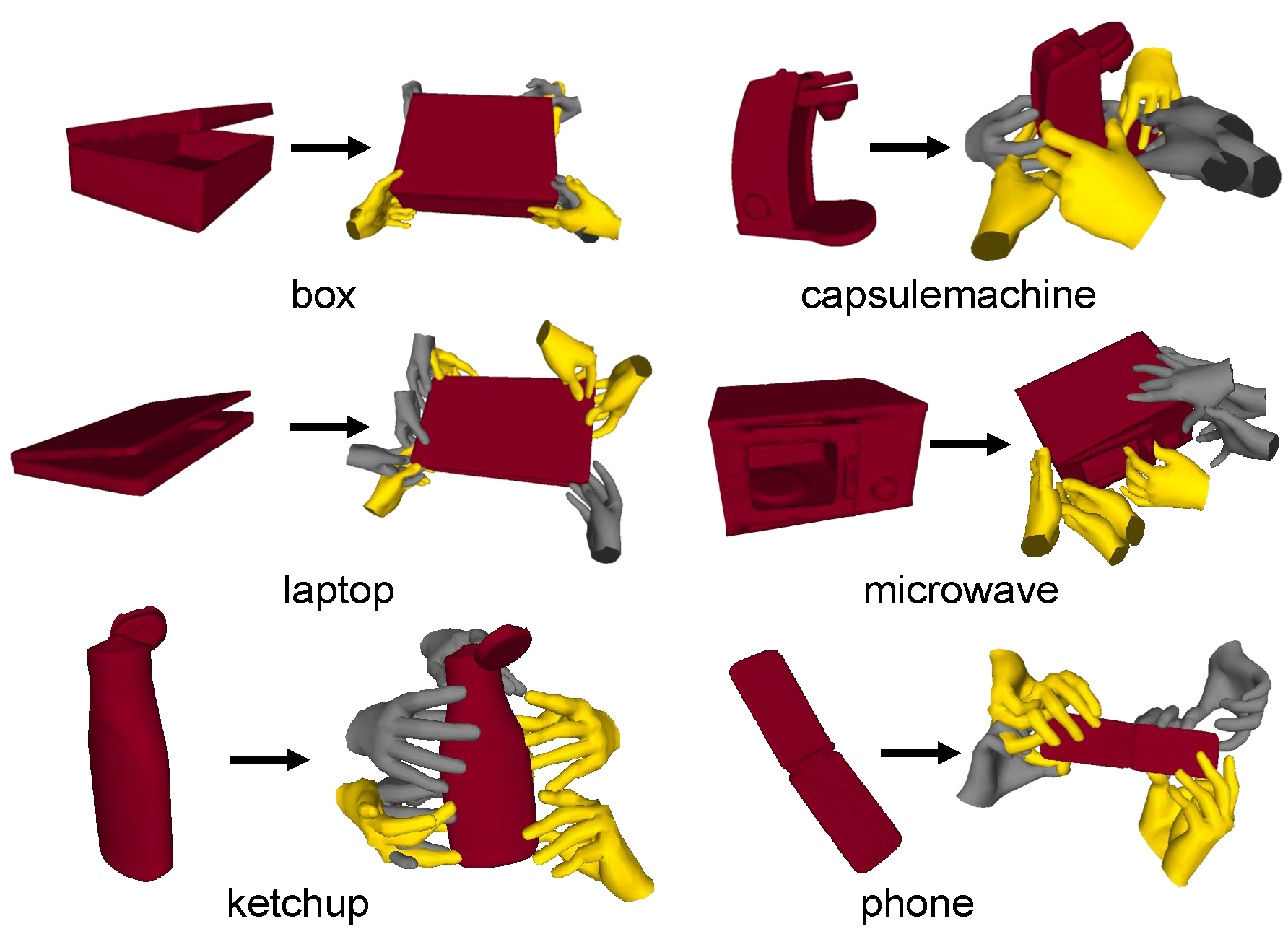}

\caption{The diversity of interactive poses after our POM optimization.
}
\label{fig:supp_pom_res}
\end{figure}

%% file: supp_pose.tex
\begin{figure*}[ht]
\centering%
\includegraphics[width=\linewidth]{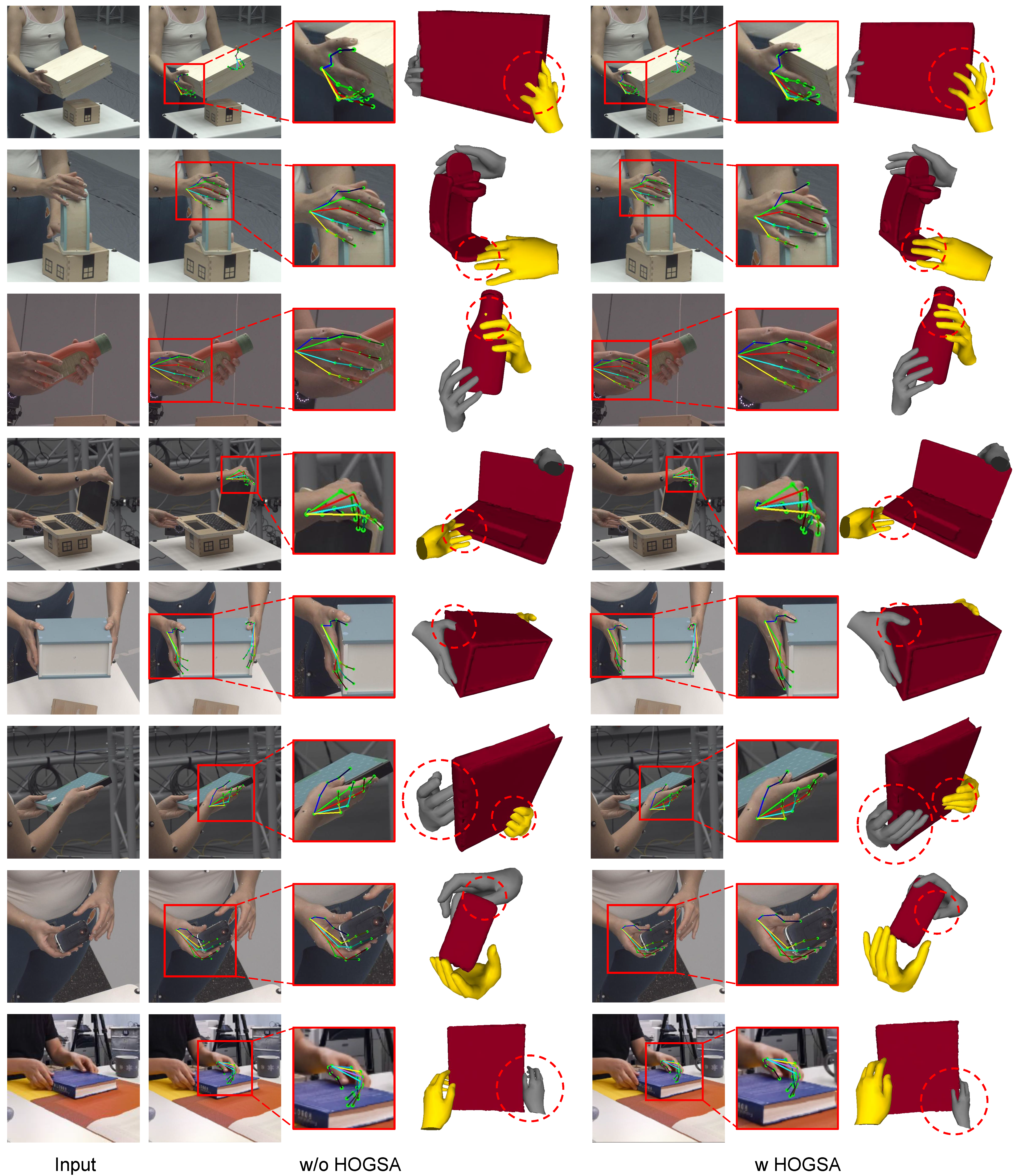}

\caption{Comparison of consistent motion reconstruction. After optimizing the baseline using our data augmentation, the model can predict more accurate results.
}
\label{fig:supp_pose}
\end{figure*}

%% file: supp_srm.tex
\begin{figure*}[ht]
\centering%
\includegraphics[width=\linewidth]{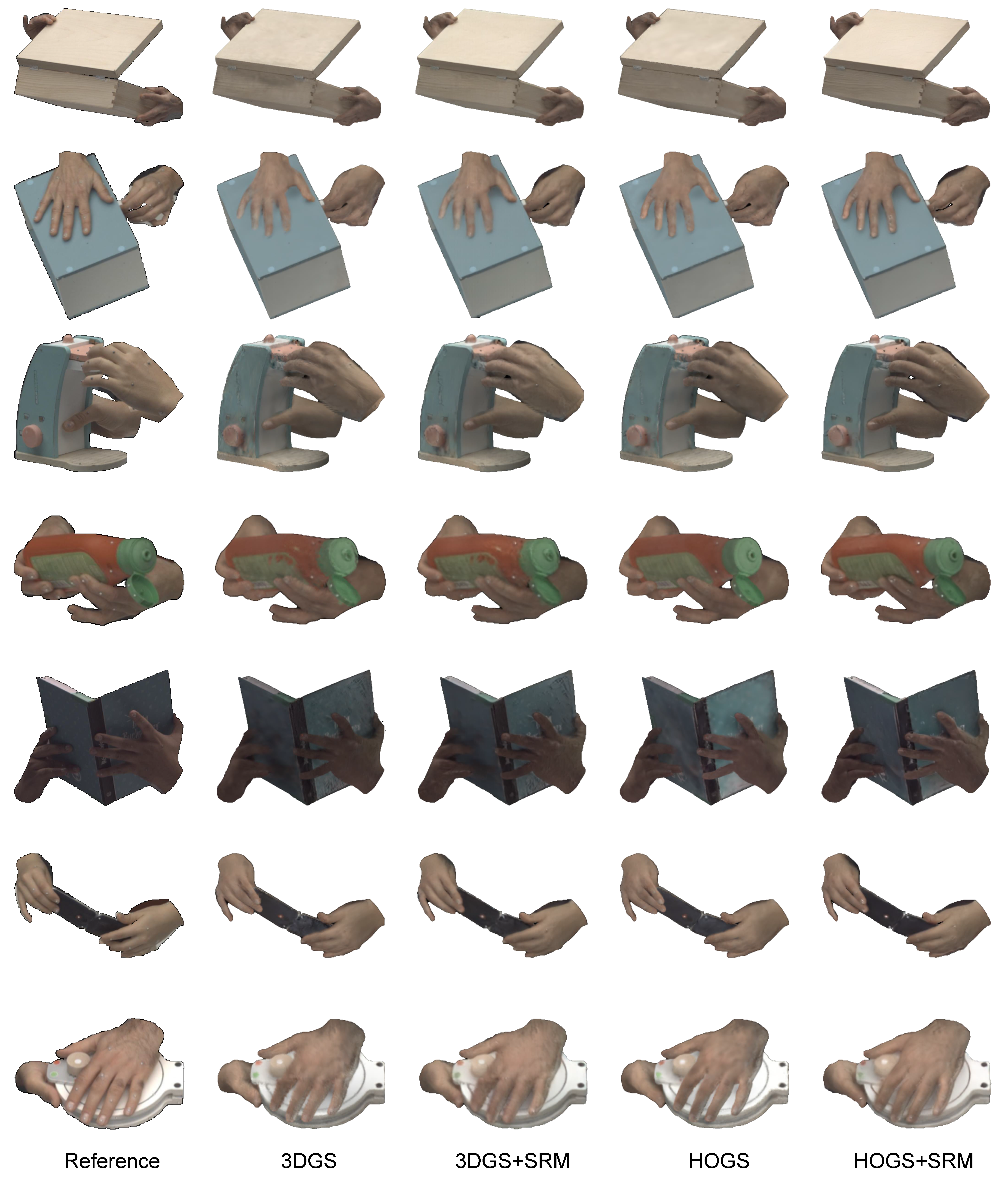}

\caption{Comparison of rendering quality. Our mesh-based method captures more realistic textures, and our SRM can be applied to different 3DGS methods to further improve rendering quality.
}
\label{fig:supp_srm}
\end{figure*}

%% file: supp_app.tex
\begin{figure*}[ht]
\centering%
\includegraphics[width=0.95\linewidth]{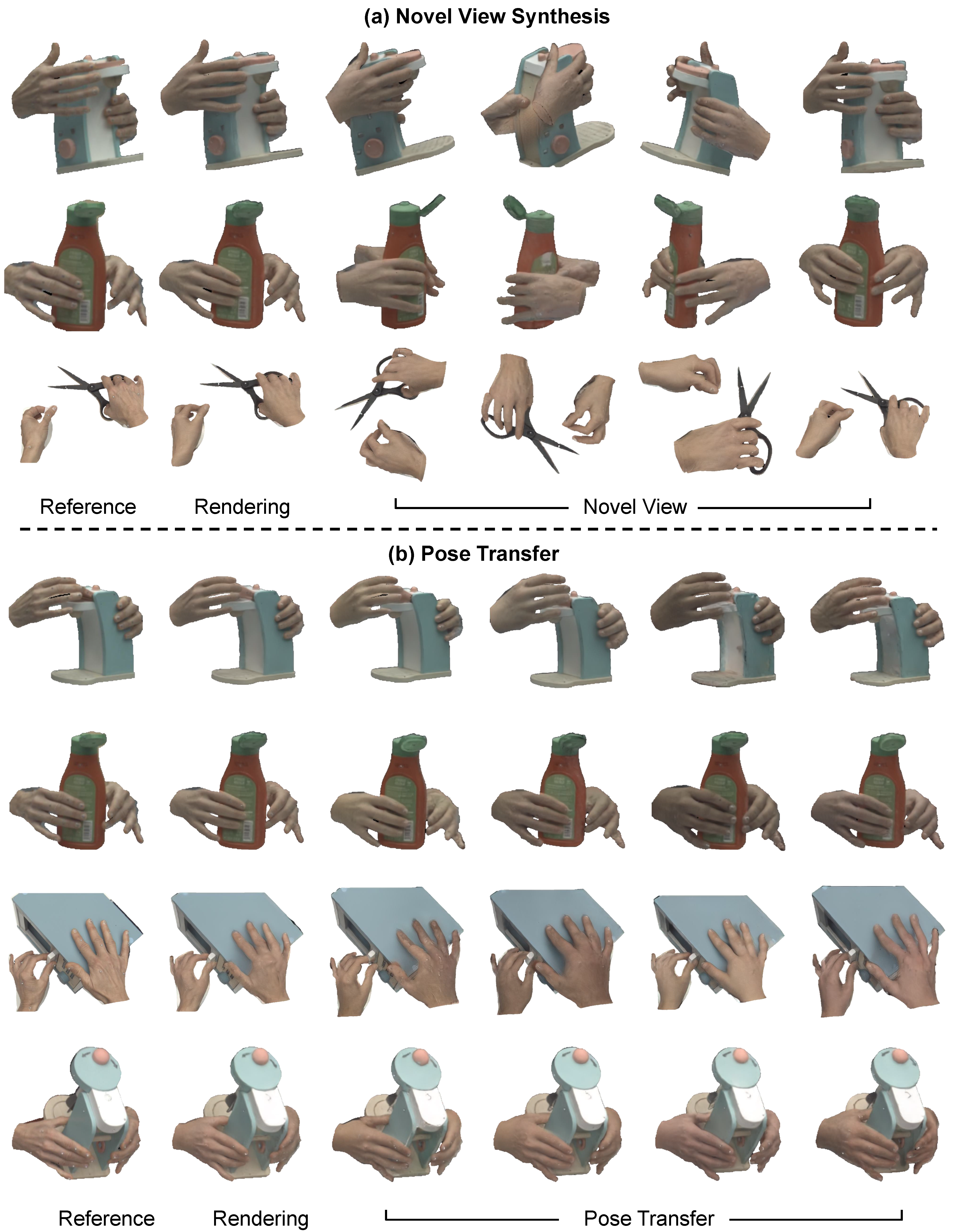}

\caption{Our method can support applications such as novel view synthesis and pose transfer.
}
\label{fig:supp_app}
\end{figure*}